\documentclass{article}

\usepackage{PRIMEarxiv}

\usepackage[utf8]{inputenc} 
\usepackage[T1]{fontenc}    
\usepackage{hyperref}       
\usepackage{url}            
\usepackage{booktabs}       
\usepackage{amsfonts}       
\usepackage{nicefrac}       
\usepackage{microtype}      
\usepackage{lipsum}
\usepackage{fancyhdr}       
\usepackage{graphicx}       
\graphicspath{{media/}}     

\usepackage{amsmath,amsfonts}
\usepackage{algorithmic}
\usepackage{graphicx}
\usepackage{textcomp}
\usepackage{xcolor}
\usepackage{amsmath}
\usepackage{todonotes}
\usepackage{pifont}

\usepackage{times}

\usepackage{soul}
\usepackage{url}

\usepackage{amsmath}
\usepackage{bm}
\usepackage{color}
\usepackage{pgfplots}
\usepackage{subcaption}
\usepackage{wrapfig, blindtext}
\usepackage{comment}
\usepackage{pgfplotstable}

\usepackage{balance}
\usepackage{algorithmic}
\usepackage[ruled,vlined,linesnumbered]{algorithm2e}

\definecolor{darkgreen}{rgb}{0, 0.0, 0}
\definecolor{darkblue}{rgb}{0, 0.0, 0.0}
\definecolor{darkred}{rgb}{0.0, 0.0, 0.0}

\title{Solving Dynamic Graph Problems with Multi-Attention Deep Reinforcement Learning}

\author{
Udesh Gunarathna\\
University of Melbourne\\
Melbourne\\
Australia\\
pgunarathna@student.unimelb.edu.au\\
\And
Renata Borovica-Gajic\\
University of Melbourne\\
Melbourne\\
Australia\\
renata.borovica@unimelb.edu.au\\
\And
Shanika Karunasekara\\
University of Melbourne\\
Melbourne\\
Australia\\
karus@unimelb.edu.au\\
\And
Egemen Tanin\\
University of Melbourne\\
Melbourne\\
Australia\\
etanin@unimelb.edu.au\\
}

\begin{document}
\maketitle

\begin{abstract}
Graph problems such as traveling salesman problem, or finding minimal Steiner trees are widely studied and used in data engineering and computer science. Typically, in real-world applications, the features of the graph tend to change over time, thus, finding a solution to the problem becomes challenging. The dynamic version of many graph problems are the key for a plethora of real-world problems in transportation, telecommunication, and social networks. In recent years, using deep learning techniques to find heuristic solutions for NP-hard graph combinatorial problems has gained much interest as these learned heuristics can find near-optimal solutions efficiently. However, most of the existing methods for learning heuristics focus on static graph problems. The dynamic nature makes NP-hard graph problems much more challenging to learn, and the existing methods fail to find reasonable solutions.

In this paper, we propose a novel architecture named Graph Temporal Attention with Reinforcement Learning (GTA-RL) to learn heuristic solutions for graph-based dynamic combinatorial optimization problems. The GTA-RL architecture consists of an encoder capable of embedding temporal features of a combinatorial problem instance and a decoder capable of dynamically focusing on the embedded features to find a solution to a given combinatorial problem instance. We then extend our architecture to learn heuristics for the real-time version of combinatorial optimization problems where all input features of a problem are not \emph{known a priori}, but rather learned in real-time. Our experimental results against several state-of-the-art learning-based algorithms and optimal solvers demonstrate that our approach outperforms the state-of-the-art learning-based approaches in terms of effectiveness and optimal solvers in terms of efficiency on dynamic and real-time graph combinatorial optimization.
\end{abstract}


\section{Introduction}
Graph combinatorial problems have been studied by computer scientists for decades as these problems appear in many fields, including transportation, computer networks, social networks, and logistics planning. Most of the existing approaches focus on the static version of the problems where the input attributes do not change over time. Unfortunately, many real-world applications of these problems in transportation, telecommunication, and social networks require the dynamic version to be addressed where the input attributes change over time \cite{dynamic-tsp, dynamic_vrp, icde_predict1}. 
    
Let us consider a motivating example for a dynamic graph problem, a ride sharing scenario, where multiple customers need to be picked up from several locations in a road network, and each customer needs to be dropped off at different locations. The objective (i.e., the optimization goal) is to find an order to pick up and drop off customers providing the fastest possible travel time. This is an NP-hard optimization problem, and can be reduced to a minimum Steiner tree \cite{ride-share}. In real-world road networks, the travel time between pickup and drop-off locations changes over time due to congestion levels or other external factors such as accidents. Thus, while optimizing the order of visiting customers, we need to consider the changes in the travel time between different locations; the road network graph is dynamic. 
Many road network related graph problems  \cite{dynamic_lane,dynamic_survey,dynamic_ta} approximate this case and assume that even though the travel times between locations are not static, changes in travel times are known beforehand (typically by relying on historical data or through traffic predictions \cite{icde_predict1,icde_predict2}). This common problem then becomes a dynamic graph combinatorial optimization problem.
If we assume that the future travel time estimates are not available and the future travel times between locations are only available after we reach that time, then the problem becomes real-time graph combinatorial optimization. 
A similar scenario can happen in telecommunication networks where we need to allocate bandwidths to communication links to maximize the overall network utilization while traffic in each link changes over time\cite{icde_link_util}. With these real-world examples, it is evident that the dynamicity of graph attributes need to be considered to make them applicable for many real-world scenarios. 
    
    
Due to the NP-hard nature, even in static combinatorial optimizations, finding an exact solution for a large instance is computationally expensive \cite{np-hard}. On the other hand, heuristic solutions can be fast but require domain knowledge of a given problem and a significant amount of manual-engineering while designing the heuristics \cite{nips-s2v}. 
    Finding such a heuristic in a dynamic setting can be even more challenging due to a large number of possible combinations\footnote{For example, Travelling Salesmen Problem (TSP) with $n$ number of nodes has $n!$ possible selections. In dynamic TSP, since each node has different values at different time steps (assuming it takes $n$  time steps to traverse), the possible combinations would be $(n!)^2 = (n.n \times (n-1)(n-1) ...$).}. In this work we thus seek a generalized method to find fast and efficient heuristics for dynamic graph problems without relying on  hand-crafted methods. 
    
    The recent advancements in Deep Neural Networks (DNN) and Reinforcement Learning (RL) have led to finding heuristic solutions for combinatorial optimization efficiently without labor intensive  hand-crafted engineering \cite{nips-s2v,aaai-eco-dqn,lclr-attn,cpaior-attn,nips-vpr}. All these approaches model combinatorial optimization as a graph (complete or incomplete) and formulate the decision-making process as a successive addition of nodes to the solution. By doing so, the learned heuristics can achieve near-optimal performance. However, most of these works focus only on static data~\cite{lclr-attn,cpaior-attn,nips-vpr}. As we demonstrate in our experiments, these algorithms fail to find reasonable solutions in dynamic graph combinatorial optimization problems.
    
    To identify the shortcomings of the mentioned approaches for dynamic settings, we categorize the existing methods into two main types: (1) Message passing neural networks with value-based reinforcement learning~\cite{nips-s2v,aaai-eco-dqn}, and (2) Attention-based/Recurrent Neural Networks (RNN) with policy-gradient reinforcement learning~\cite{lclr-attn,cpaior-attn,nips-vpr}. The former methods use a message-passing neural network to iteratively encode the graph information and use Q-learning \cite{q-learning} for the decision making. The latter use an attention mechanism~\cite{nips-nlp} or an RNN to encode the graph instance as a sequence of nodes and use a decoder to make the decisions iteratively. As recent research suggest~\cite{lclr-attn,nips-vpr}, the latter approaches achieve better performance for combinatorial optimizations largely due to the fact that the policy-gradient RL algorithms perform better when the action space (i.e., the number of nodes to select in a combinatorial problem) is large/variable~\cite{sutton}. 
    
   Despite the success of attention-based policy gradient methods in static settings, the reason for them not performing well in dynamic combinatorial optimizations is three-fold. (1) The graph/input features are embedded before making decisions/actions. Thus, changes in graph/input features due to the decision-making process are not captured, which is the case in dynamic combinatorial problems. (2) The attention mechanism deployed is only able to encode graph node features in a single time step. In a dynamic setting, each node contains different features at different time steps. The above-mentioned attention-based models are unable to capture such temporal changes. (3) The decoding mechanism used by the attention-based methods focuses on the entire embedded space given by the encoder at every decision-making time step. However, in a dynamic setting, there are different encoded outputs at every time step. Thus, paying attention to the entire embedded space tends to distract the decision-making process in dynamic settings.
 
     We propose a novel encoder-decoder architecture named \emph{Graph Temporal Attention (GTA)} trained with modified \emph{Reinforcement Learning (RL)} to address the aforementioned issues and we denote \emph{GTA-RL} as the combined overall architecture. First, we propose a multi-dimensional attention mechanism at the encoder to embed both temporal and graph (spatial) features simultaneously. Then, a fusion mechanism is introduced to learn the inter-relationship between temporal and spatial embeddings. Such encoder layers are cascaded together. Once the set of encoder layers outputs the spatio-temporal representation of the dynamic CO, a novel decoder named temporal pointing decoder is used to dynamically pay attention to the spatio-temporal representation for the decision-making stage. Finally, GTA-RL is trained through a modified policy-gradient RL algorithm. Note that, this architecture targets the scenarios where all the dynamic changes are estimated beforehand as in the standard dynamic CO problems \cite{dynamic-tsp,tsc-max-flow}. Next, we introduce an iterative method by partially embedding the encoder inside the decoder to tackle the problem where the dynamic changes are not \emph{known a priori}. Thus, making GTA-RL applicable for a variety of real-time applications.

    The contributions from this work are four-fold:
    \begin{itemize}
        \item We propose a novel deep learning architecture, GTA-RL, to tackle the dynamic graph combinatorial optimization problems.
        \item We introduce an encoder and decoder that are capable of preserving temporal information in sequential decision making.
        \item We propose an iterative encoder-decoder architecture that can handle real-time information of a graph combinatorial optimization problem.
        \item Our extensive experimental results demonstrate that GTA-RL achieves superior performance compared to state-of-the-art learning-based algorithms in the associated combinatorial optimization domain as demonstrated on an important class of graph problems.
    \end{itemize}
    
\section{Related Work} \label{sec-related}

Until recently, the success of Deep Neural Networks (DNN) was mostly limited to prediction tasks \cite{lclr-attn}. With the advancements in reinforcement learning algorithms combined with DNNs emerged Deep Reinforcement Learning (DRL) algorithms, which demonstrated great success in sequential decision making and control problems in many domains including atari games \cite{nature-atari}, traffic control \cite{ecml-drl,aaai-drl}, and robot navigation \cite{icra-drl}. In recent years, DRL has been applied to many combinatorial optimization problems to learn heuristics that require no human input~\cite{aaai-eco-dqn,cpaior-attn,nips-s2v,nips-vpr,lclr-attn,arx-a2c}. These works have achieved near-optimal performance outperforming hand-crafted heuristics in many combinatorial optimization problems. These existing methods formulate the combinatorial graph problems as a successive addition of graph nodes (actions) to the solution. The existing methods can be broadly categorized into two types, based on the reinforcement algorithm adopted in the solutions. (1) Using Value-based methods such as Deep Q-learning~\cite{nature-atari}. (2) Using Policy-gradient algorithms such as REINFORCE and Actor-Critic algorithms~\cite{sutton}. 

Value-based method for graph combinatorial optimization was first proposed by Dai \emph{et al.}\cite{nips-s2v}. The proposed architecture uses \emph{structured2vec} to embed the graph information and uses fitted Q-learning \cite{ecml-q-fitted} for sequential decision making. They applied the proposed solution to solve static Travelling Salesmen (TSP), Minimum Vertex Cover, and Max-Cut problems. A similar architecture is followed by Barrett \emph{et al.}\cite{aaai-eco-dqn} using Message Passing Neural Networks. However, in their approach, the solution is further optimized at inference time, which improves over a one-shot solution in Dai \emph{et al.}\cite{nips-s2v}.

The policy-gradient methods have demonstrated improved results over value-based based counterparts, as evident by the recent research \cite{cpaior-attn,lclr-attn,arx-a2c}. The main reason is because in graph combinatorial optimization, the number of nodes (the action space) can be variable from a problem instance to instance and when the action space is large, the policy-gradient algorithms tend to perform better \cite{sutton}. The first policy-gradient algorithm for graph combinatorial optimization was proposed by Bello \emph{et al.}\cite{arx-a2c}. The paper uses an LSTM encoder-decoder architecture named \emph{pointer network} which was first proposed by Vinyals \emph{et al.}\cite{nips-pn} to solve the TSP in a supervised learning setting. However, Bello \emph{et al.} uses policy-gradient instead of supervised learning to train the pointer network. In addition, Bello \emph{et al.} improves the decoder by a masking scheme to mask already selected nodes. Nazari \emph{et al.} \cite{nips-vpr} uses a different \emph{pointer network} architecture employing an attention mechanism and RNN to solve the Vehicle Routing Problem (VRP). Their solution includes a separate dynamic encoder to embed the dynamic features of a given combinatorial optimization problem. Despite being able to handle dynamic changes to the problem over time, this architecture is developed mainly focusing on the internal changes due to the node selection in the decision-making process but not the changes due to external factors which are beyond the control of the decision making process. 
As our experiments show, GTA-RL achieves superior performance in several combinatorial problems over the architecture of Nazari \emph{et al.}~\cite{nips-vpr}. 

The next generation of policy-gradient methods uses pure attention architectures instead of RNNs inspired by the architecture proposed in Vaswani \emph{et al.}\cite{nips-nlp}. Deudon \emph{et al.} uses a Multi-head attention \cite{nips-nlp} for encoding the combinatorial optimization problem and a decoder with a pointing mechanism similar to Bello \emph{et al.} but without recurrent elements. The REINFORCE algorithm \cite{sutton} with a critic baseline has been used for training the network. In parallel to Deudon \emph{et al.}, Kool \emph{et al.} \cite{lclr-attn} proposed an improved architecture by introducing a new attention-based decoder instead of the pointing mechanism. The paper also shows that using a rollout baseline instead of the critic baseline improves the results further for combinatorial optimization problems. Kool \emph{et al.} outperforms existing methods in TSP, VRP (Vehicle Routing Problem), and variants of both TSP, VRP. \textcolor{darkgreen}{Peng \emph{et al.} \cite{AM-D} extends this architecture to the VRP problem where changes from an agent behaviour are taken into account. However, external changes to the graph problem are not considered. A similar graph attention network approach is followed by Drori \emph{et a.l}\cite{icmla-gat} as well.} Our proposed architecture GTA-RL uses an architecture akin to Kool \emph{et al.}, and extend it with a novel encoder and decoder, which are capable of embedding temporal features of dynamic combinatorial optimization problems and dynamically focusing on temporal features during the decision making process. Our experimental results show that we outperform the afore-mentioned state-of-the-art approaches for dynamic combinatorial optimization problems. We also extend our architecture to real-time combinatorial optimization. 

We provide a summary of the existing methods in Table \ref{tbl:comp}.

\begin{table*}[]
\centering
\begin{tabular}{|c|c|c|c|c|c|}
\hline
Algorithm                                & Decision making algorithm & Type of neural network    & Static & Dynamic & Real-time \\ \hline
Dai \emph{et al.}\cite{nips-s2v}       & Value-based                                                         & Message-passing                                                      & \checkmark      & -       & \checkmark          \\ \hline
Barrett \emph{et al.}\cite{aaai-eco-dqn} & Value-based                                                         & Message-passing                                                      & \checkmark       & -       & \checkmark          \\ \hline
Bello \emph{et al.}\cite{arx-a2c}      & Policy-gradient                                                     & RNN                                                                  & \checkmark       & -       & -         \\ \hline
Vinyals \emph{et al.}\cite{nips-pn}    & Supervised learning                                                       & RNN                                                                  & \checkmark       & -       & -         \\ \hline
Nazari \emph{et al.}\cite{nips-vpr}                    & Policy-gradient                                                     & RNN                                                                  & \checkmark       & -       & -         \\ \hline
Deudon \emph{et al.}\cite{cpaior-attn}                    & Policy-gradient                                                     & Pure-attention                                                       & \checkmark       & -       & -         \\ \hline
Kool \emph{et al.} \cite{lclr-attn}    & Policy-gradient                                                     & Pure-attention                                                       & \checkmark       & -       & -         \\ \hline
Our approach (GTA-RL)                    & Policy-gradient                                                     & Spatio-temporal attention & \checkmark       & \checkmark        & \checkmark          \\ \hline
\end{tabular}
\caption{The table shows a summary of existing methods and GTA-RL. The \checkmark indicates whether a particular algorithm supports a feature.}
\label{tbl:comp}

\end{table*}

\section{Dynamic Graph Combinatorial Optimization as a Learning Problem} \label{sec-prob-def}

In this section, we present a generic dynamic graph combinatorial problem formulation and discuss the approach to parameterize the problem formulation so that it can be learned.

First, we represent the dynamic graph combinatorial problem as a graph instance $G$ with $N$ number of nodes. The graph features can be represented as $X = \{x_1, x_2, ..., x_N\}$, \textcolor{darkblue}{where for each node $i$, $x_i = \{x_{0,i}, x_{1,i}, ..., x_{T-1,i}\}$} is a vector with the dimension $\mathbb{R}^{T \times D}$. $T$ represents the total number of time-steps and $D$ is the number of features in one input element at a given time step. We denote $x_{i,t} \in \mathbb{R}^D$ as the input features of node $i$ at time $t$. The cost of selecting node $j$ at time step $t$, after selecting node $i$ at the previous time step, is represented as a \emph{cost function} $f_c: x_{i,t} \times x_{j,t} \mapsto \mathbb{R}$. In a complete graph, the cost function can have a value from any node $i$ to any node $j$ at any given time step $t$. In an incomplete graph, the cost function for two nodes where there is no edge can be ignored by setting a large negative value for such connections (a.k.a. masking). Note that the complete graph scenario can also be considered as a simple sequence of nodes without any graph/edge properties, resulting in dynamic combinatorial optimization.  

Given the above details, our objective is to find an order of nodes $Y$ with length $T_s \in (0, T]$ which satisfies the constraints of a given dynamic graph combinatorial optimization $C(Y)$, such that we minimize:

\begin{equation} \label{eqn-obj}
    P_{obj}[Y|G] = \sum_{t=1}^{T_s} f_c(x(y_t,t),x(y_{t+1}, t+1))
\end{equation}

\textcolor{darkblue}{where $y_t \in Y$ is a selected node from $G$ for time step $t$.} The $C$ is an evaluating function that checks whether the sequence $Y$ satisfies all the problem constraints\footnote{For example, in dynamic TSP, the cost function, $f_c$, represents the distance between two nodes at a given time step $t$.  The evaluating function, $C$, checks whether all nodes have been reached or not. The objective, $P_{obj}$, represents the total traveled distance after reaching every node.}.

\textbf{Learning Formulation: } Now that we formulated the dynamic graph combinatorial problem, our objective is to find a policy, $\pi$. which will generate a sequence $Y$ such that we minimize $P_{obj}$ and satisfy $C(Y)$, for a given problem instance $G$. First, we can parameterize the policy as $\pi_{\theta}(Y|G) := Pr(Y|G)$. Next, we can factorize this as a Markov Decision Process (MDP) where input states $S$ will be the graph instance $G$, plus the solution computed up to the given time step $t$ ($S=\{G,Y_{1:t-1}\}$) and the action $y_t$ is selected from available nodes in $G$. Then, we factorize $\pi_{\theta}(Y|G)$ using the probability chain rule as:

\begin{equation} \label{eqn-dyn-factorized}
    \pi_{\theta}(Y|G) = \Pi_{t=1}^{T}\pi_{\theta}(y_t|G,Y_{1:t-1})
\end{equation}

Finally, our objective transforms to learn and represent the set of parameters $\theta$ in Equation \ref{eqn-dyn-factorized}, so that we minimize $P_{obj}$.

\textbf{Real-time Dynamic Graph Combinatorial Optimization: } Here, we formulate the real-time graph combinatorial optimization problem where the input attributes of a given problem $G$ at time step $t$ are not available until we reach that time step. Unlike the dynamic graph combinatorial optimization, now we do not have all the information about the graph for the entire time horizon. In that case, Equation \ref{eqn-dyn-factorized} can be written as below.

\begin{equation} \label{eqn-rlt-factorized}
    \pi_{\theta}(Y|G) = \Pi_{t=1}^{T}\pi_{\theta}(y_t|G_{1:t-1},Y_{1:t-1})
\end{equation}

Note that, in Equation \ref{eqn-rlt-factorized}, the graph instance is $G_{1:t-1}$ which indicates that only the information up to time step $t$ is available. Equation \ref{eqn-rlt-factorized} is harder to optimize than the dynamic combinatorial optimization version since we are acting with incomplete information. However, we later show that with a slight modification to our proposed architecture, we can solve the real-time combinatorial optimization to be on par with dynamic combinatorial optimization. 

\textbf{Remark:} The real-time combinatorial problems typically occur in transportation networks where the road network is represented as a graph. The edge attributes can be traffic loads or average speeds of a road segment. Given that we want to reach several points in the road network, first, we decide on our first point to reach based on the current edge attributes. Then, we take a finite amount of time to reach that point. During that time, the edge attributes may have changed due to external traffic conditions. However, these external traffic information could not have been accessed beforehand. This is a typical example of a real-time combinatorial problem. 

\section{Graph Temporal Attention with Reinforcement Learning}

To find a solution to the dynamic combinatorial optimization problem defined in Equation \ref{eqn-dyn-factorized}, we need to find a neural network architecture to represent $\theta$ parameters and a learning algorithm to train the neural network. We propose an encoder-decoder neural network architecture named \emph{Graph Temporal Attention (GTA)} to represent $\theta$ parameters. Then, we use a loss function computed through policy-based reinforcement learning to train $\theta$ parameters. This section provides details about these two components. 

\subsection{Graph Temporal Attention}

A major limitation of the previously proposed attention based models \cite{lclr-attn, cpaior-attn} was the inability to  encode the time-varying features of a problem and make a decision while input features are changing. To address these limitations, GTA network follows an encoder-decoder architecture consisting of two components, namely a \emph{temporal encoder} and a \emph{temporally pointing decoder}.

\begin{figure}
    \centering
    \includegraphics[width=5.8cm]{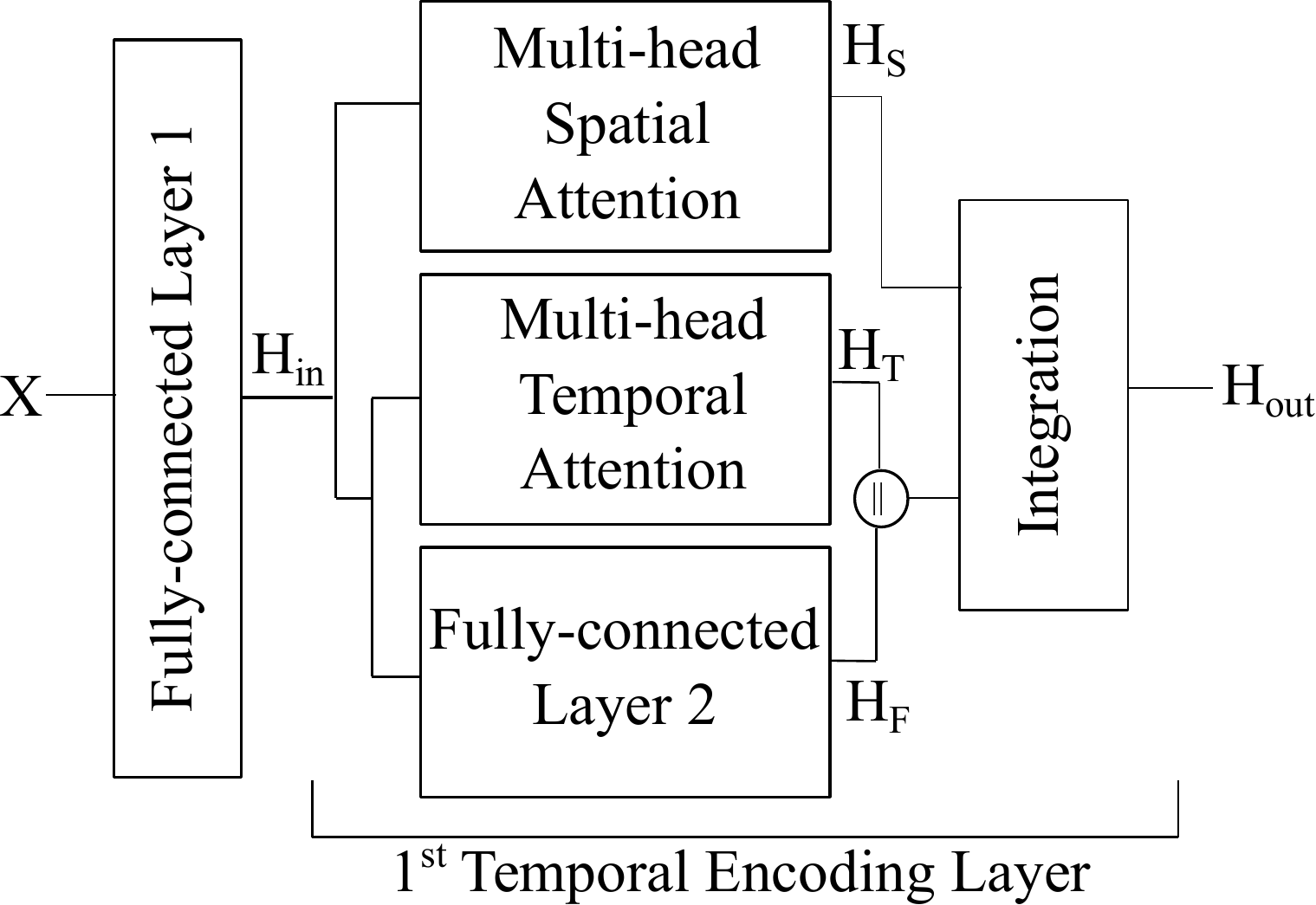}
    \caption{Temporal Encoder with one layer. The input features $X$ will be encoded with an initial \emph{Fully-connected Layer 1} before the first temporal encoding block. The temporal encoder's three parallel layers are shown next named \emph{Spatial Attention, Temporal Attention, Fully-connected Layer 2}. The outputs from these layers are combined at the \emph{Integration Layer} and the final output is provided to the next temporal encoding layer.}
    \label{fig:temp-encoder}
\end{figure}

\subsubsection{Temporal Encoder}

Figure \ref{fig:temp-encoder} depicts the temporal encoder architecture. First, input $X \in \mathbf{R}^{T \times N \times D}$ is transformed by a linear transformation to $H^{(0)} \in \mathbf{R}^{T\times N\times D^h}$, at \emph{fully connected layer 1}. $D^h$ refers to the hidden dimension of \emph{fully connected layer 1}. The output from \emph{fully connected layer 1} is given to the temporal enoder. The temporal encoder contains a spatial attention, a temporal attention and a fully-connected layer in parallel. $H^{(0)}$ is given as an input to these three layers and the outputs from these layers ($H_S^{(1)}, H_T^{(1)}, H_F^{(1)}$) are combined through an integration layer. For simplicity, we only show the first encoding layer in Figure \ref{fig:temp-encoder}, however, several temporal encoders can be stacked by taking the output of first temporal layer $H^{(1)}$ as the input to the second layer, and so on. We denote $H^{(l)}$ as the input to the $l^{th}$ temporal encoding layer and $H^{(l+1)}$ as the output from that layer. We denote $h_{i,t}^{(l)}$ as the hidden representation of node $i$ at time $t$ in layer $l$.

\textcolor{darkgreen}{Even though the embedding information along a temporal axis has not been investigated in the field of learning heuristics for combinatorial optimization, this idea has recently emerged in other prediction tasks \cite{aaai-gman,st-lstm,aaai-lstm}. Zheng \emph{et. al} \cite{aaai-gman} uses a neural architecture named \emph{STAtt Block} which consists of two attention mechanisms and a fixed graph embedding for traffic prediction, while Gangopadhyay \emph{et al.} \cite{st-lstm}, and Pareja \emph{et al} \cite{aaai-lstm} use an LSTM-based mechanism. Compared to LSTM, attention-based methods are more computationally efficient and have the ability to handle graph based information \cite{nips-nlp}. Therefore, our proposed temporal encoder,} whilst motivated by \emph{STAtt Block}, differs in several key aspects. First, in our temporal encoder we do not use the fixed graph embedding. This is because in traffic prediction tasks learning happens for a single road network topology, whereas in combinatorial optimizations the input graph topology changes for every training instance. Thus, it is infeasible to learn a separate embedding for every new training instance. Second, in dynamic combinatorial problems, there can be some nodes in $X$ whose attributes do not change over time. On the contrary, in traffic prediction attributes such as traffic load or congestion of every node change over time. In the temporal encoder, we introduce the \emph{fully connected layer 2} in parallel to \emph{temporal attention layer}, to handle nodes whose attributes do not change over time.

Next, we detail each sub-element in the temporal encoder.

\vspace{0.0cm}

\textbf{Spatial Attention Layer:} In a given graph instance, there are dependencies between nodes, and the spatial attention layer is used to encode those dependencies. 

The spatial attention layer uses the self-attention mechanism introduced by Vaswani {et al.}\cite{nips-nlp}. The spatial attention layer considers a list of node features at a given time step as a sequence: $H_{S,t}^{(l)} = \{h_{1,t}^{(l)}, h_{2,t}^{(l)}, ... h_{N,t}^{(l)}\} \in \mathbb{R}^{N \times D^h}$. The spatial attention uses the same standard parameters named query ($wS_{q}^{(l)}\in\mathbb{R}^{D^h \times D^k}$), key ($wS_{q}^{(l)}\in\mathbb{R}^{D^h \times D^k}$) and value ($wS_{q}^{(l)} \in \mathbb{R}^{D^h \times D^v}$). In this setup, we set $D^k$ and $D^v$ to the same dimension as $D^h$. However, the derivations of the paper can be applied even when these values are different. 

First, we compute the following three projections as below.

\begin{align*}
    & qS_t^{(l)} = wS_{q}^{(l)}H_{S,t}^{(l)} \in \mathbb{R}^{N \times D^h} \\
    & kS_t^{(l)} = wS_{k}^{(l)}H_{S,t}^{(l)} \in \mathbb{R}^{N \times D^h} \\
    & vS_t^{(l)} = wS_{v}^{(l)}H_{S,t}^{(l)} \in \mathbb{R}^{N \times D^h} 
\end{align*}

Then, we compute the scaled dot product between the query and the key projections and use a \emph{softmax layer} as in Vaswani {et al.}\cite{nips-nlp}.

\begin{equation} \label{eqn-sp-weights}
    \alpha_t = softmax(\frac{(qS_t^{(l)})^T.kS_t^{(l)}}{\sqrt{D^h}}) \in \mathbb{R}^{N \times N}
\end{equation}

Here, $\alpha_{t,i,j} \in \alpha_{t}$, represents the relevance of node $j$ to node $i$ at time $t$. In a non-complete graph, we can set $-\infty$ to non-adjacent nodes before computing the softmax value.  Finally, we can output the spatial attention for each node with their relevance to others as:

\begin{equation} \label{eqn-sp-value}
    H_{S,t}^{(l+1)} = \alpha_t.vS_t^{(l)} \in \mathbb{R}^{N \times D^h}
\end{equation}

By stacking $H_{S,t}^{(l+1)}$ over the time axis, we get the $H_S^{(l+1)}$ as below.

\begin{equation}
    H_S^{(l+1)} = \mathbin\Vert_{t=1}^T H_{S,t}^{(l+1)} \in \mathbb{R}^{T \times N \times D^h}
\end{equation}

\begin{figure*}[t]
    \centering
    \includegraphics[width=15cm, height=4.6cm]{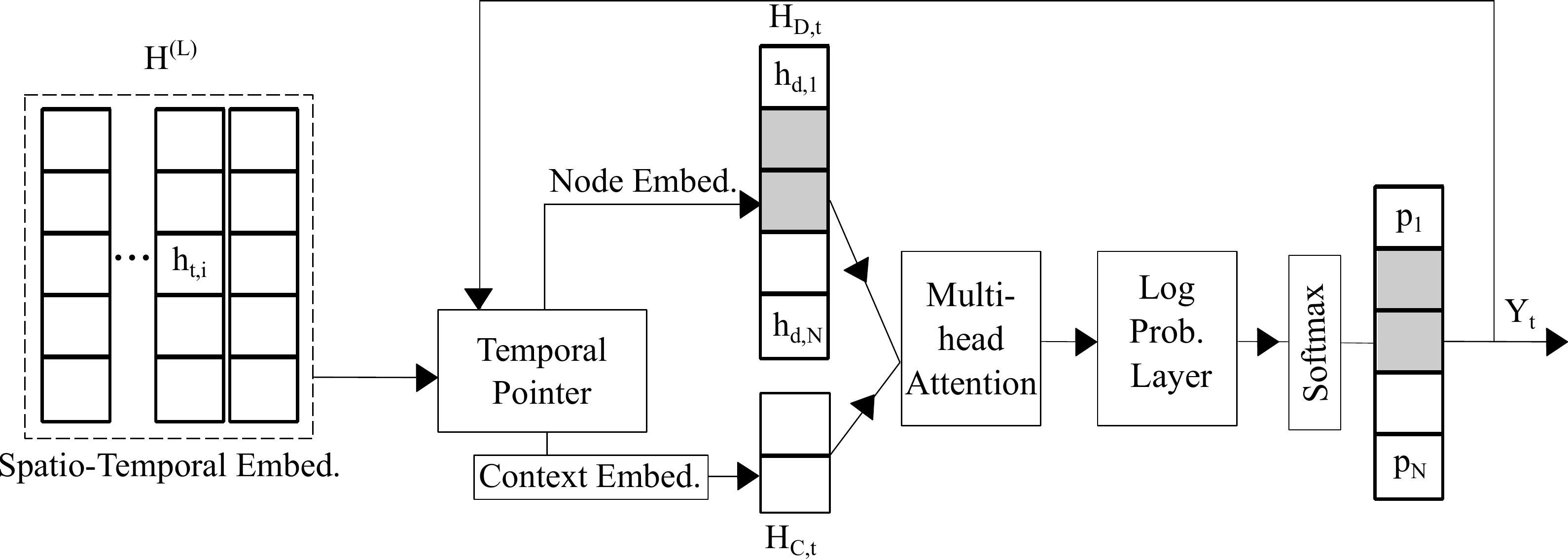}
    \caption{This figure depicts the entire decision making process of the Temporal Decoder. The temporal decoder consists of a \emph{Temporal Pointer} which adaptively points to the part of the encoded output from the temporal encoder(\emph{Node Embed}). The temporal decoder then generates a \emph{Context. Embed} which represents the context of the problem, given its partial solution. These two outputs will be fused together at \emph{Multi-head Attention Layer} and the final probabilities for each node will be computed at \emph{Log Prob. Layer}. The node with the highest probability is selected as a part of the solution and will be added to $Y_t$. During the entire process, invalid nodes will be ignored.}

    \label{fig:temp-decoder}
\end{figure*}
    
\vspace{0.0cm}
\textbf{Temporal Attention Layer with fully Connected Layer:} In dynamic combinatorial problems, in addition to the dependency between nodes, there is a dependency between node features across different time steps for the same node. To encode such dependencies, the temporal attention layer is introduced. 

We also note that in some dynamic combinatorial problems, apart from nodes that change over time, there can be some nodes that are not changing over time as well. For example, in \emph{Vehicle Routing Problem} defined in Section \ref{sec-problems}, the features of the depot node (vehicle starting/loading node) may be fixed while customer node features change over time. Feeding such fixed nodes' information into the same temporal attention layer will reduce the effectiveness of the learning of dependencies of nodes that \emph{are} changing over time.

To resolve this issue, we introduce a separate fully connected layer. First, we note that there can be $U$ number of such fixed nodes and these nodes are represented by the set $\mathbb{U}$. Then, from input $H^{(l)}$, we take $H_F^{(l)} \in \mathbb{R}^{T, U,D^h} $ which only contains the nodes with fixed features. In the fully connected layer, we apply a linear transformation for $H_F^{(l)}$ to get $H_F^{(l+1)}$. 

Then, in the temporal attention layer, we select the nodes that are changing over time and  represent them as $H_T^{(l)} \in \mathbb{R}^{T \times  N-U \times D^h}$. Similar to the spatial attention layer, we consider a sequence, but now in temporal dimension for each node $i$ as $H_{T,i}^{(l)} = \{h_{i,1}^{(l)}, h_{i,2}^{(l)}, ... h_{i,T}^{(l)}\} \in \mathbb{R}^{T \times D^h}$, where node $i \notin \mathbb{U}$. The temporal attention uses the same standard parameters named query ($wT_{q}^{(l)}\in\mathbb{R}^{D^h \times D^h}$), key ($wT_{q}^{(l)}\in\mathbb{R}^{D^h \times D^h}$) and value ($wT_{q}^{(l)} \in \mathbb{R}^{D^h \times D^h}$).

First, we compute the following three projections as below:

\begin{align*}
    & qT_i^{(l)} = wT_{q}^{(l)}H_{T,i}^{(l)} \in \mathbb{R}^{T \times D^h} \\
    & kT_i^{(l)} = wT_{k}^{(l)}H_{T,i}^{(l)} \in \mathbb{R}^{T \times D^h} \\
    & vT_i^{(l)} = wT_{v}^{(l)}H_{T,i}^{(l)} \in \mathbb{R}^{T \times D^h} 
\end{align*}

Then, we compute the scaled dot product between the query and the key projections like in the spatial attention layer.

\begin{equation} \label{eqn-tp-weights}
    \beta_i = softmax(\frac{(qT_i^{(l)})^T.kT_i^{(l)}}{\sqrt{D^h}}) \in \mathbb{R}^{T \times T}
\end{equation}

Here, $\beta_{i,t_a,t_b} \in \beta_{i}$, represents the relevance of the node features at time $t_b$ to the node features at $t_a$ for node $i$. In contrast to the spatial attention, we do not use the masking and assume all time steps are connected. The temporal attention layer outputs $H_{T,i}^{(l+1)}$, for each node as below.

\begin{equation}\label{eqn-sp-value}
    H_{T,i}^{(l+1)} = \beta_t.vS_i^{(l)} \in \mathbb{R}^{T \times D^h}
\end{equation}

By stacking $H_{T,i}^{(l+1)}$ for nodes that are changing over time, we compute $H_T^{(l+1)}$ as below.

\begin{equation}
    H_T^{(l+1)} = \mathbin\Vert_{i=1,i \notin \mathbb{U}}^N H_{T,i}^{(l+1)}
\end{equation}

Finally, the outputs from the temporal attention layer and the fully connected layer are concatenated to get the final temporal embedding representation. 

\begin{equation}
 H_{TF}^{(l+1)} = H_T^{(l+1)} \mathbin\Vert H_F^{(l+1)} \in \mathbb{R}^{N \times T \times D^h}
\end{equation}

\textbf{Remark:} The factorization of the temporal attention layer with the fully connected layer can be extended to multiple parallel layers as well. Suppose one knows how the nodes evolve over time according to some underlying distributions, then more than one temporal attention layer and more than one fully connected layer can be applied. Of course, how one may gain such knowledge on the existence of different distributions and the decision on the number of layers is beyond the scope of this paper and may be investigated in future research. 

\textbf{Multi-head Attention:} As suggested by Vaswani {et al.}\cite{nips-nlp}, using more than one attention head (a.k.a. multi-head attention) improves the stabilization of the learning process. In multi-head attention, $D^k$ and $D^v$ defined above will be equal to $D^h/M$, where $M$ is the number of attention heads. The final output will contain $M$ number of embeddings with a dimension of $\mathbb{R}^{N \times D^h/M}$ and these $M$ embeddings can be concatenated or projected through another weight vector to get a representation of shape $\mathbb{R}^{N \times D^h}$. 
Both the temporal and spatial attention layers use the multi-head attention mechanism. 

\vspace{0.0cm}

\textbf{Integration Layer:} Once we learned the spatial representation ($H_S^{(l+1)}$) and the temporal representation ($H_{TF}^{(l+1)}$), we need to learn how to combine these two representations.

First, we concatenate both $H_S^{(l+1)}$ and transposed $H_{TF}^{(l+1)}$ which results in a shape of $\mathbb{R}^{T \times N \times 2*D^h}$. Then, we use a linear transformation with a weight vector $wI \in \mathbb{R}^{D^h \times 2*D^h}$ and use a sigmoid activation layer.

\begin{equation}
 H^{(l+1)} = \sigma(wI.(H_S^{(l+1)} \mathbin\Vert H_{TF}^{(l+1)})) \in \mathbb{R}^{T \times N \times D^h}
\end{equation}

Finally, $H^{(l+1)}$ is given to the next layer $l+1$ in the temporal encoder. Our integration layer is much simpler compared to the layer proposed in \emph{STAtt Block}~\cite{aaai-gman}, but yields similar performance.

The integration layer output is designed in a way that for a given node $i$ at time $t$, integration layer accounts for both the impact from surrounding nodes as well the previous and the future time variants of node $i$.

\subsubsection{Temporally Pointing Decoder} \label{sec-decoder}

This section describes the decision making process of the decoding layer of GTA-RL, which takes as input the embedded information from the encoder output. In literature, the decoding layers in static combinatorial optimization problems have used \emph{pointer networks}\cite{nips-vpr} or a combination of multi-head and single-head attention \cite{lclr-attn} to select the next node in a solution. Since the combination of multi-head and single-head attention has traditionally yielded better results and better convergence rate over the \emph{pointer networks}\cite{lclr-attn}, we  follow a similar multi-head and single-head attention architecture, and extend it with components that handle temporal features. 

Figure \ref{fig:temp-decoder} shows the overall architecture of the temporally pointing decoder. The temporally pointing decoder works in a sequential manner with a feedback loop. First, it takes the spatio-temporal output from the last layer ($L$) of the encoder ($H^{(L)}$) and the decoded solution up to the current time step $t$. Then, the temporal pointer outputs a node embedding representation highlighting the features of the current time step. The invalid nodes for the selection will be ignored during the computation of node embedding representation. This masking is problem-dependent\footnote{For example, in dynamic TSP, already visited nodes from previous time steps are invalid nodes.}. Third, the temporal pointer also outputs a context embedding by considering the current decoded solution and the current time step. These two outputs will be then fused together in a multi-head attention layer. Then, a single attention layer named \emph{Log Probability Layer} is used to find the probability of each node being in the current solution. The node with the highest probability is selected as the next node and the output will be given as a feedback to the temporal pointer for the next iteration. The individual components are detailed next. 

\textbf{Temporal Pointer:} First, we describe the motivation behind the temporally pointing decoder. In the original multi-head and single-head attention architecture \cite{lclr-attn}, a set of fixed attention values are computed from the encoder output (where the encoder output contains a shape of $\mathbb{R}^{N \times D^h}$). These fixed attention values are used in every decoding time step. Even though the fixed attention works in a static problem, we identify that use of fixed attention is the main limitation for handling dynamic information because at every decoding time step, the node features change in dynamic combinatorial problems. Thus, initially computed fixed attention values do not reflect the node feature changes. Also, using such a fixed attention hinders the ability of the overall architecture to handle changes to the graph instance (problem) due to the new node selections. Thus, we propose to dynamically compute these attentions by taking temporal information into account.

To dynamically compute these attentions, first, let us look at the output $H^{(L)}$ of the temporal encoder. The output contains temporal dimensions, which results in a shape of $\mathbb{R}^{T \times N \times D^h}$. A naive way to handle the temporal axis is to compute a mean value for each node over the entire time horizon as below.

\begin{equation} \label{eqn-naive}
    H_{naive} = \frac{\sum_{t=1}^T H^{(L)}[t, :, :]}{T} \in \mathbb{R}^{N \times D^h}
\end{equation}

where $H^{(L)}[t, :, :]$ represents the embedding for all the nodes at time $t$. Another variant is to use only a single time-step data such as $H^{(L)}[0, :, :]$. Both of these can then be given as an input to the multi-head and single-head attention architecture to decode node selection as we demonstrate in our experiment in Section \ref{sec-variations}. However, the use of $H_{naive}$ or \ $H^{(L)}[0, :, :]$ yields similar results mainly because the resulting representation suppresses the rich information about the difference in each time step.

To alleviate this problem, we propose a way to dynamically focus on the most relevant parts of the embedded output from the encoder at every time step. We slice up the encoder embedding at every decoding time step and dynamically compute attention weights using multi-head attention and update the context of CO problem (described next). In this way, node representations are updated at every time step, and we compute the multi-head attention based on new values computed at the current time step. We achieve superior results using this technique because now we can retain information about every time step.  

\textbf{Context Embedding:} The context embedding is used to identify the context of a given problem at a decoding time step. This is obviously problem-dependent. For the context embedding, all the context embeddings are updated based on the current node representation from the encoder output $H^{(L)}$.

To generalize the context embedding to all the combinatorial problems, we define a function named context embedding function $f_{cnxt} : \mathbb{R}^{N \times T \times N \times D^h} \mapsto \mathbb{R}^{K \times D^h + e}$. The value of $K$ and $e$ will be problem dependent. $H_{C,t} = f_{cnxt}(H(L)_{t}, Y_t)$ where $H_{C,t}$ is the context embedding. Refer to Section \ref{sec-problems} for derivation of $H_{C,t}$ for different dynamic combinatorial problems. 

\textbf{Multi-head and Log Probability Layer: }
Once we have $H_{C,t}$ and $H_{D,t}$, we use the multi-head attention layer to combine the context-specific embedding with the current graph representation. Similar to the encoder, we use query $wC_q (\in \mathbb{R}^{{K*D^h + e \times D^h}}$), key $wC_k$ and value $wC_v$ weights in the attention layer. In contrast to the self-attention, $H_{C,t}$ for query and $H_{D,t}$ for both the key and value parameters is used.

\begin{align*}
    & qC_t = wC_{q}H_{C,t} \in \mathbb{R}^{N \times D^h} \\
    & kC_t = wC_{k}H_{D,t} \in \mathbb{R}^{N \times D^h} \\
    & vC_t = wC_{v}H_{D,t} \in \mathbb{R}^{N \times D^h} 
\end{align*}

Next, similar to Equation \ref{eqn-sp-weights} and Equation \ref{eqn-sp-value}, we can compute the attention weights and then the final embedding $H^{(F)}_{D,t}$.

\begin{equation}
    H^{(F)}_{D,t} = softmax\frac{(qC_t^T kC_t)}{D^h}.vC_t 
\end{equation}

The log probabilities for each node are computed using another weight vector, $wP$ and we use \emph{tanh} activation function. We clip values beyond $|C|$ similar to Bello \emph{et al.}\cite{arx-a2c}. 

\begin{equation}
    \gamma_{t} = tanh(H^{(F)}_{D,t}.(wP.H_{D,t})^T)
\end{equation}

Finally, a \emph{softmax} layer is used to compute final probabilities for each node, and the node with the highest probability is chosen as the next node and added to solution sequence $Y$.

\begin{equation}
    P_{t} = softmax(\gamma_t) \in \mathbb{R^{N}}
\end{equation}

\subsection{GTA-RL for Real-time Graph Combinatorial Problems:} The architecture we described so far assumes all input features of the problem instance are available beforehand. In real-time applications, all changes to the input features may not be available before we make the decisions. They are only available once we take the action. 

To handle this type of problems, we propose the encoder to be iterative. Next, we describe this iterative method. At the start of the problem, we only take the first time step input features, that is $x_1 \in \mathbb{R}^D$ (defined in Section \ref{sec-prob-def}). The input feature $X_1$ does not have any temporal axis as this is the node information about a single time step. First, we encode $x_1$ through the spatial encoder and send it to the decoder. The encoder also buffers the input $x_1$. Then, we modify the temporal pointer of the decoder to always points to the last time step of the embedded input features. This operation results in pointing to the most recent embeddings at every time step. Next, the decoder will select the first node for the solution sequence as the action. Then, we get the input at the next time step $x_2$. We concatenate the previous input, $x_1$, with the current input, $x_2$, and repeat the same process mentioned above. The iterative architecture follow the same approach until we find the complete solution. 

In this proposed iterative method, we do not need to take the input features for the entire time horizon. However, by buffering the previous input, the model is able to gain knowledge about the transitions of input features over time. Since it takes some time to obtain such knowledge, the model may take more time initially before converging to a satisfactory solution. Also, compared to our original implementation, the encoder now encodes the input multiple times, thus increasing the computational requirements. This is an acceptable trade-off as  real-time combinatorial problems are more difficult to optimize compared to dynamic combinatorial problems.

\subsection{Training Graph Temporal Attention with Reinforcement Learning}

As we discussed in the introduction, the policy-gradient-based methods achieve better results in combinatorial optimizations. Thus, we use, REINFORCE algorithm, a policy-gradient-based algorithm \cite{sutton} to train the Graph Temporal Attention network. First, we define the performance measure of dynamic combinatorial optimization ($G$) using objective function $P_{obj}$ (defined in Section \ref{sec-prob-def}) and using Equation \ref{eqn-dyn-factorized}.

\begin{equation}
    J(\theta|G) =  \mathop{\mathbb{E}}_{Y\sim\pi_\theta}[P_{obj}(Y|G)]
\end{equation}

Then, we can use the \emph{policy gradient theorem} to find the derivative of $J(\theta|G)$. 
\begin{equation} \label{eqn-reinforce}
    \nabla_{\theta}J(\theta|G) =  \mathop{\mathbb{E}}_{Y\sim\pi_\theta}[P_{obj}(Y|G)\nabla_{\theta}log(\pi_{\theta}(Y|G))]
\end{equation}

This derivative can be used to update parameters of GTA-RL. However, due to the high variance results in Equation \ref{eqn-reinforce}, a baseline is used to accelerate the learning. Then, Equation \ref{eqn-reinforce} can be rewritten as:

\begin{equation} \label{eqn-reinforce-baseline}
    \nabla_{\theta}J(\theta|G) =  \mathop{\mathbb{E}}_{Y\sim\pi_\theta}[(P_{obj}(Y|G)-b(G))\nabla_{\theta}log(\pi_{\theta}(Y|G))]
\end{equation}

where $b(G)$ is called the baseline function, which is a function that is independent of $Y$. 

Once we established Equation \ref{eqn-reinforce-baseline}, we need to choose a proper baseline function. There are two baselines that have been used in this context; critic baseline \cite{cpaior-attn} and rollout baseline \cite{lclr-attn}. We experimented with both these baseline functions and found that using rollout baseline works better than critic baseline for dynamic combinatorial optimization. 

\section{Experimental Setup}

We evaluate GTA-RL on two routing-based combinatorial optimization problems named \emph{Travelling Salesman Problem (TSP)} and \emph{Vehicle Routing Problem (VRP)}. In this section, we first describe these two problems, then we discuss our experimental environment, and the parameters and the baselines used for our experimental evaluation. 

\subsection{Problems} \label{sec-problems}
\textbf{Dynamic Travelling Salesman Problem: } Similar to the static combinatorial problem, there are $n$ number of points (nodes) in the euclidean space, and the objective is to find the order of nodes to visit to cover all the nodes such that the total traveled distance is minimal. In dynamic TSP, the node locations are changing uniformly at random in euclidean space, i.e., x,y coordinates are changing uniformly at random at every time step\footnote{The initial node locations are assigned uniformly at random between 0 and 1 in 2d-space. Then, at every time step, the node locations are updated uniformly at random with maximum change of 0.1 in coordinates.}. In this way, the cost of traveling between two nodes are changing over time in the euclidean space. This is similar to the travel time changes in a transportation network due to external traffic loads. Also, changing node's x,y coordinates means we are updating the node features. Therefore, this scenario can be easily generalized to any problem where node features change over time such as in telecommunication networks or social networks. 

In previous works~\cite{nips-s2v,lclr-attn,cpaior-attn}, for static combinatorial problems, training and testing were done using networks with 20, 50, and 100 nodes. However, due to the added computational complexity in dynamic problems, a similar number of possible solutions is generated by a smaller number of nodes. To illustrate, static TSP with 100 nodes contains $100! (=9.332622e+157)$ possible solutions and dynamic TSP with 50 nodes contains $(50!)^2 (=9.250171e+128)$ possible solutions, i.e., yields the same complexity with half the number of nodes. Thus, in this paper, we test dynamic TSP with 10, 20, 50 nodes, and denote them as TSP10, TSP20, and TSP50, respectively. For sensitivity analysis experiments in Section \ref{sec-variations}, we select 20 as the default value. 

We introduced a problem-dependent context embedding $H_C$ in Section \ref{sec-decoder} under \textbf{Context Embedding} subsection. For travelling salesmen problem, $H_C$ contains 3 elements; the first selected node, the last selected node and the graph embedding computed by summing up all nodes at time $t$ as below.

\begin{equation} \label{eqn-tsp-context}
  \text{TSP: } H_{C,t} = \{h^{(L)}_{y_0} || h^{(L)}_{y_t} || H_{G,t}\} \text{;  } H_{G,t} = \sum_{i=0}^Nh^{(L)}_{t,i}
\end{equation}

\textbf{Vehicle Routing Problem:} The vehicle routing problem is one of the most challenging combinatorial optimization problems. The conventional vehicle routing problem contains a set of $n$ number of points in euclidean space and a vehicle with some capacity $c$. Each node $i$ has demand $d_i$ of goods to satisfy and $d_i < c$ for all the nodes. Out of these $n$ nodes, one node is called a depot, and the vehicle can visit the depot and fill goods until the capacity. When visiting a node other than the depot, the vehicle should have enough goods to satisfy the demand of that particular node. Similar to the previous TSP, the node coordinates are updated uniformly at random at every time step. However, in VRP, we do not change the location of the depot to demonstrate a node that is not changing over time. Finally, the objective is to find the minimum distance the vehicle needs to travel to satisfy the demands of all the nodes. Similar to TSP, we run experiments with VRP consisting of 10, 20, and 50 nodes. These are denoted as VRP10, VRP20, and VRP50, respectively.

For the vehicle routing problem, the context embedding is the last selected node embedding plus the remaining capacity of the vehicle, and the graph embedding is the same as in TSP.

\begin{equation} \label{eqn-vrp-context}
  \text{VRP: } H_{C,t} = \{ h^{(L)}_{y_t} || r || H_{G,t}\} 
\end{equation}

where $r$ is the remaining capacity of the vehicle.

\textbf{Real-time Versions: } We use the same formulation in real-time TSP similar to dynamic TSP; however, only the node locations at that time step is provided which is similar to the real-world ride-sharing scenario described in the introduction where the future traffic congestion information is not available. Once a node is selected, we reach that node and the next node to visit is selected, thus in a real-time setting order of the visited nodes cannot be updated even if we found a better order of nodes in a later time step.
The same principle applies to real-time VRP as well.

\subsection{Parameter Settings}
We now detail the hyper-parameters used in our experimental setup. In GTA-RL, we use three layers of the temporal encoder, and the hidden dimension $D^h$ is set to 128. We use 12800 problem instances during one epoch, and the batch is set to 32 instances. A total of 50 epochs is used for the training. We use a learning rate of $1e^{-4}$ for both the actor and the critic of the GTA-RL. We use Adam optimizer \cite{adam} for the training of the GTA-RL. The experiments were conducted in a machine consisting of an Intel Xeon(R) processor with 24 GB RAM and an Nvidia GRID P40 GPU. The hyper-parameters are the same for both VRP and TSP training. Our code base is built on top of the following code base\footnote{\url{https://github.com/wouterkool/attention-learn-to-route}} using Pytorch and can be accessed here\footnote{\url{https://github.com/udeshmg/GTA-RL}}. 

\subsection{Baselines}

We use several baselines for the comparison. The baselines include learning heuristics, hand-crafted heuristics, and optimal solutions computed through Integer Programming\footnote{
Note that we are not using the algorithm proposed by Barrett \emph{et al.}\cite{aaai-eco-dqn} in our comparison because their implementation only supports Max-cut problem, but not for TSP or VRP. }. 

\begin{itemize}
    \item \textbf{S2V-DQN:} This solution was proposed by Dai \emph{et al. }\cite{nips-s2v}. S2V-DQN uses structure2vec for graph encoding and uses fitted Q-learning \cite{ecml-q-fitted} for the decision making, which belongs to value-based methods as described in Section \ref{sec-related}. S2V-DQN is implemented in a way that a node in the problem instance can be selected only once. Therefore, S2V-DQN cannot be used for VRP, since VRP depot node is visited multiple times. In TSP, there are two variants of S2V-DQN depending on how the reward function is formulated. The first method is to provide a negative reward for increased distance by adding the last selected node to the end of the tour of TSP (S2V-DQN-last). The second is to provide a negative reward for increased distance by adding the last selected to the middle of the tour where the resulting distance is minimum (S2V-DQN-sorted). S2V-DQN-sorted improves the solution quality significantly, but we should point out that S2V-DQN-sorted cannot be used in a real-time setting where the tour cannot be changed once selected. 
    
    \item \textbf{RNN-RL:} This solution was proposed by Nazari \emph{et al.}\cite{nips-vpr}. RNN-RL uses two recurrent networks with REINFORCE algorithm and belongs to a class of policy-gradient base algorithms. RNN-RL contains two encoders named static and dynamic. In the original implementation coordinates of nodes were provided for the static encoder. We provided coordinates of nodes to the dynamic encoder as the coordinates are no longer static and change over time.

    \item \textbf{AM:} This solution was proposed by Koot \emph{et al.}\cite{lclr-attn}. The architecture is limited to handle a fixed set of nodes, and we use the initial locations of nodes as an input to the AM model for dynamic TSP and VRP cases. 
    \textcolor{darkgreen}{\item \textbf{AM-D:} This solution is an extension of \textbf{AM}, with a dynamic encoder used to solve VRP \cite{AM-D}.}
    
    \item \textbf{Gurobi:} We implement integer programming solutions for both dynamic TSP and dynamic VRP using the Gurobi\footnote{https://www.gurobi.com} python library. The integer programming approach is able to find the optimal solutions for both of the problems. Computational times for dynamic TSP and VRP are however significantly higher than their static counterparts. For instance, 900 seconds for dynamic TSP size 20 vs around 2.5 seconds for static TSP with 20 nodes. Thus, we were only able to compute gurobi optimization for TSP20 and VRP10. We should also note that specialized solvers such as concorde\footnote{https://www.math.uwaterloo.ca/tsp/concorde.html} cannot be used in our comparisons as these do not support dynamic combinatorial optimizations. 
    
    \item \textbf{Nearest Neighbor(NN):} We also implement a hand-crafted heuristic named Nearest Neighbor(NN) for dynamic TSP problem. Here, the next node to visit at the next time step is selected based on the distance to surrounding nodes, favouring closest nodes first. The original NN is developed for static TSP, and we extend it to a version of NN that can handle the dynamic TSP. In extended NN, we compute the nearest neighbors for a given node at the current time step using the coordinates of the neighboring nodes at the next time step.
\end{itemize}

\subsection{GTA-RL Variations}

There are several variations of GTA-RL that we use for the evaluation. We use this section to name these variations formally.
\begin{itemize}
    \item \textbf{GTA-RL-greedy: } This is the standard GTA-RL, and it uses the greedy decoding in the decoder, where the node with the highest probability is selected as the next node.
    
    \item \textbf{GTA-RL-bs: }. Unlike GTA-RL-greedy, GTA-RL-bs uses the \emph{beam search} technique during the decoding, where nodes with ($k$) number of highest probabilities are kept at every time step and decoding in the next time step is done with respect to previously-stored $k$ values. Finally, the sequence of nodes with the highest probability according to Equation \ref{eqn-dyn-factorized} is selected. We should also note that GTA-RL-bs cannot be used in a real-time setting as the order of nodes changes.
    
    \item \textbf{GTA-RL-sum} We briefly discussed a naive implementation of temporal decoder in Section \ref{sec-decoder}. The method was to sum up the encoding over the time axis of the embedding given by the encoder. GTA-RL-sum uses this technique. 
    
    \item \textbf{GTA-RL-(0)} Similar to GTA-RL-sum, in this algorithm we only use the first time step of the embedding given by the encoder. 
    
    \item \textbf{GTA-RL-rt} This is the version of GTA-RL that is capable of handling real-time data. 
    
\end{itemize}

\pgfplotsset{select coords between index/.style 2 args={
    x filter/.code={
        \ifnum\coordindex<#1\def\pgfmathresult{}\fi
        \ifnum\coordindex>#2\def\pgfmathresult{}\fi
    }
}}

\pgfplotstableread[col sep = comma]{Tex/DataTSP.csv}\datatsp
\pgfplotstableread[col sep = comma]{Tex/DataVRP1.csv}\datavrp

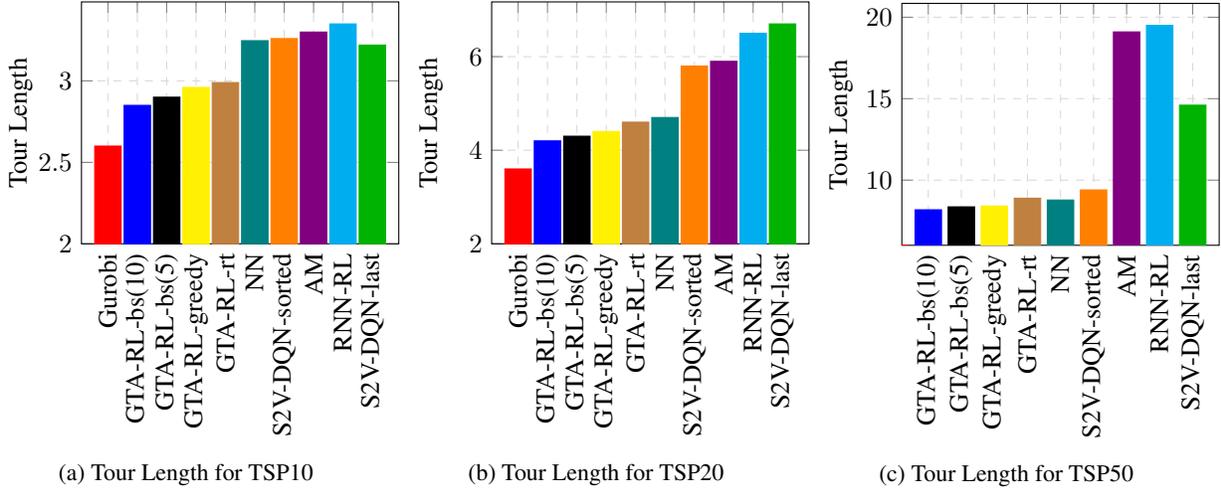
\begin{figure*}[t]
\begin{subfigure}{0.30\columnwidth}
\begin{tikzpicture}
 \begin{axis}[
width=5.8cm, height=4.8cm,     
grid=major, 
grid style={dashed,gray!30}, 
x tick label style={rotate=90},
ylabel=Tour Length,
ymin=2,
ylabel style={at={(0.1,0.5)}},
legend style={at={(0.5,-0.2)},anchor=north},
cycle list name=color list,
xtick={1,2,3,4,5,6,7,8,9,10},
xticklabels={Gurobi, GTA-RL-bs(10), GTA-RL-bs(5), GTA-RL-greedy, GTA-RL-rt, NN, S2V-DQN-sorted, AM, RNN-RL, S2V-DQN-last}
]

\foreach \x in {0,..., 10}
    \addplot+[select coords between index={\x}{\x} , ybar, fill] table[x=index, y=TSP10, col sep=comma]{\datatsp};

\end{axis}
\end{tikzpicture}

\caption{Tour Length for TSP10}
\label{fig:tsp10}
\end{subfigure}\hspace{5mm}\begin{subfigure}{0.30\columnwidth}
\begin{tikzpicture}
 \begin{axis}[
width=5.8cm, height=4.8cm,     
grid=major, 
grid style={dashed,gray!30}, 
x tick label style={rotate=90},
ylabel=Tour Length,
ymin=2,
ylabel style={at={(0.1,0.5)}},
legend style={at={(0.5,-0.2)},anchor=north},
cycle list name=color list,
xtick={1,2,3,4,5,6,7,8,9,10},
xticklabels={Gurobi, GTA-RL-bs(10), GTA-RL-bs(5), GTA-RL-greedy, GTA-RL-rt, NN, S2V-DQN-sorted, AM, RNN-RL, S2V-DQN-last}
]

\foreach \x in {0,..., 10}
    \addplot+[select coords between index={\x}{\x} , ybar, fill] table[x=index, y=TSP20, col sep=comma]{\datatsp};

\end{axis}
\end{tikzpicture}

\caption{Tour Length for TSP20}
\label{fig:tsp20}
\end{subfigure}\hspace{5mm}\begin{subfigure}{0.30\columnwidth}
\begin{tikzpicture}
 \begin{axis}[
width=5.8cm, height=4.8cm,     
grid=major, 
grid style={dashed,gray!30}, 
x tick label style={rotate=90},
ylabel=Tour Length,
ymin=6,
ylabel style={at={(0.1,0.5)}},
legend style={at={(0.5,-0.2)},anchor=north},
cycle list name=color list,
xtick={1,2,3,4,5,6,7,8,9,10},
xticklabels={Gurobi, GTA-RL-bs(10), GTA-RL-bs(5), GTA-RL-greedy, GTA-RL-rt, NN, S2V-DQN-sorted, AM, RNN-RL, S2V-DQN-last}
]

\foreach \x in {0,..., 10}
    \addplot+[select coords between index={\x}{\x} , ybar, fill] table[x=index, y=TSP50, col sep=comma]{\datatsp};

\end{axis}
\end{tikzpicture}

\caption{Tour Length for TSP50}
\label{fig:tsp50}
\end{subfigure}
\caption{Tour length for dynamic TSP}

\end{figure*}

\begin{figure*}[t]
\begin{subfigure}{0.30\columnwidth}
\begin{tikzpicture}
 \begin{axis}[
width=5.8cm, height=4.8cm,     
grid=major, 
grid style={dashed,gray!30}, 
x tick label style={rotate=90},
ylabel=Tour Length,
ymin=3,
ylabel style={at={(0.1,0.5)}},
legend style={at={(0.5,-0.2)},anchor=north},
cycle list name=color list,
xtick={1,2,3,4,5,6,7,8},
xticklabels={Gurobi, GTA-RL-bs(10), GTA-RL-bs(5), GTA-RL-greedy, GTA-RL-rt, AM-D, AM, RNN-RL}
]

\foreach \x in {0,..., 8}
    \addplot+[select coords between index={\x}{\x} , ybar, fill] table[x=index, y=VRP10, col sep=comma]{\datavrp};

\end{axis}
\end{tikzpicture}

\caption{Tour Length for VRP10}
\label{fig:vrp10}
\end{subfigure}\hspace{5mm}\begin{subfigure}{0.30\columnwidth}
\begin{tikzpicture}
 \begin{axis}[
width=5.8cm, height=4.8cm,     
grid=major, 
grid style={dashed,gray!30}, 
x tick label style={rotate=90},
ylabel=Tour Length,
ylabel style={at={(0.1,0.5)}},
legend style={at={(0.5,-0.2)},anchor=north},
cycle list name=color list,
xtick={0,1,2,3,4,5,6,7,8},
ymin=6.5,
xticklabels={wild, Gurobi, GTA-RL-bs(10), GTA-RL-bs(5), GTA-RL-greedy, GTA-RL-rt, AM-D, AM, RNN-RL}
]

\foreach \x in {0,..., 7}
    \addplot+[select coords between index={\x}{\x} , ybar, fill] table[x=index, y=VRP20, col sep=comma]{\datavrp};

\end{axis}
\end{tikzpicture}

\caption{Tour Length for VRP20}
\label{fig:vrp20}
\end{subfigure}\hspace{5mm}\begin{subfigure}{0.30\columnwidth}
\begin{tikzpicture}
 \begin{axis}[
width=5.8cm, height=4.8cm,     
grid=major, 
grid style={dashed,gray!30}, 
x tick label style={rotate=90},
ylabel=Tour Length,
ylabel style={at={(0.1,0.5)}},
legend style={at={(0.5,-0.2)},anchor=north},
cycle list name=color list,
ymin=12,
xtick={0,1,2,3,4,5,6,7,8},
xticklabels={wild, Gurobi, GTA-RL-bs(10), GTA-RL-bs(5), GTA-RL-greedy, GTA-RL-rt, AM-D, AM, RNN-RL}
]

\foreach \x in {0,..., 8}
    \addplot+[select coords between index={\x}{\x} , ybar, fill] table[x=index, y=VRP50, col sep=comma]{\datavrp};

\end{axis}
\end{tikzpicture}

\caption{Tour Length for VRP50}
\label{fig:vrp50}
\end{subfigure}
\caption{Tour length for dynamic VRP}
\vspace{-4mm}
\end{figure*}
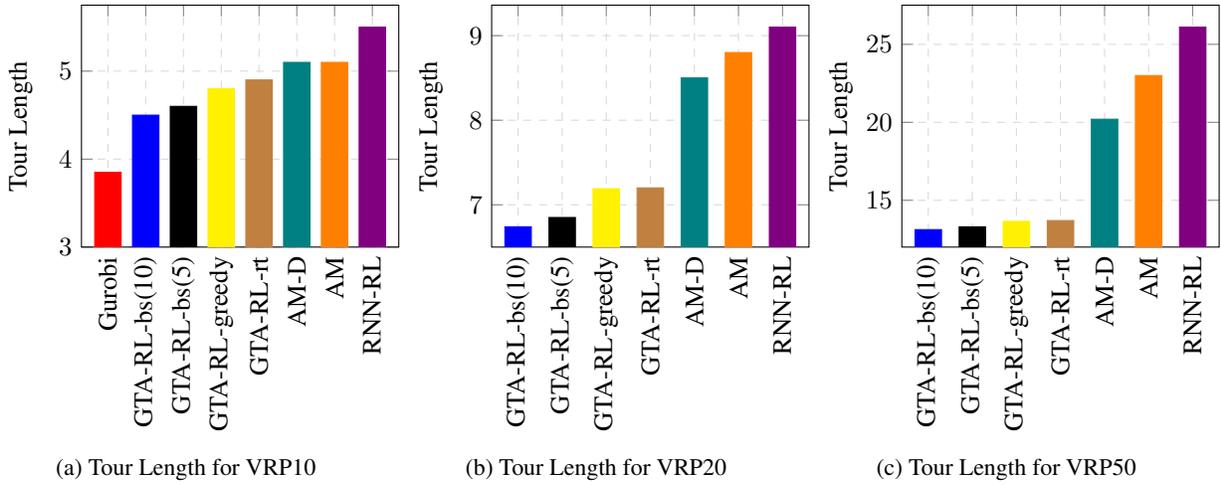

\section{Experimental Results}

 We now compare the results of GTA-RL-variations against the baselines. First, we present the results against the baselines for dynamic TSP and dynamic VRP, and then we present results for real-time TSP and real-time VRP. Following that, we compare GTA-RL-variations to reason about our architecture. Finally, we present the solutions achieved by GTA-RL compared to the optimal solution computed by Gurobi.

\subsection{Comparison for Dynamic TSP and Dynamic VRP}

\textbf{Dynamic TSP20:} Figure \ref{fig:tsp20} shows the average tour length for the validation data set achieved by each algorithm with 20 nodes (TSP20). The learning-based algorithms, S2V-DQN-last, RNN-RL, AM, are not able to find a satisfactory solution as the average tour length from these algorithms is between 5.8 to 6.3. The results demonstrate that these algorithms are unsuitable for dynamic combinatorial optimization despite them being applicable to the static settings. S2V-DQN-sorted however finds a much lower average tour length compared to S2V-DQN-last. This is because S2V-DQN-sorted is able to change the order of the selection process; once it sees the dynamic changes, S2V-DQN-sorted can shuffle the order of nodes to reduce the tour length. This shuffle however is impossible in a real-time setting where the selected node cannot be updated. The non-learning-based solution, NN, is able to find a reasonable solution as the modified NN can handle the dynamic nature of combinatorial optimization.

GTA-RL-greedy beats all the other baseline algorithms, including hand-crafted heuristic (NN). In GTA-RL, the action selection is based on the highest probability at that time step. However, employing a beam search strategy, GTA-RL-bs based algorithm finds a lower average tour length. This reduction is expected as with beam search we select the final solution by looking at $k$ ($k$ is the beam width) final possible  outcomes. The beamwidth of 10, GTA-RL-bs(10), achieves the lowest average tour length\footnote{Note however that the use of beam width for real-time applications is not possible, since GTA-RL-bs(10) needs to find the whole tour before selecting the best possible tour.}
We should also highlight that the GTA-RL-greedy solution is not far from GTA-RL-bs(10), indicating that we can produce a high-quality solution even with the greedy strategy. 

One could argue that GTA-RL can have an unfair advantage over the value-based approaches such as S2V-DQN-last. This is because the S2V-DQN also works in a real-time fashion where future inputs are not taken into account. The rationale is justified with the results of GTA-RL-rt. GTA-RL-rt considers the problem as a real-time problem and does not take future inputs while computing the solution. Thus, GTA-RL-rt and S2V-DQN-last work in the same setting. However, as depicted in Figure \ref{fig:tsp20}, GTA-RL-rt is able to find a much shorter tour length compared to S2V-DQN-last. This highlights the effectiveness and flexibility of our approach.

By comparing the optimal controller, Gurobi, with the best learning-based algorithm from Figure \ref{fig:tsp20}, the results show that GTA-RL-bs(10) is quite close to the optimal solution, with only 10\% gap between the Gurobi optimal solution's average tour length and GTA-RL-bs(10). As explained before, we believe that the complexity of dynamic combinatorial problems makes it much harder to find a heuristic solution compared to the static scenario, making 10\% gap acceptable. Also, our objective is not to beat an optimal optimizer such as Gurobi, but to propose a heuristic for \emph{efficiently} solving the dynamic combinatorial optimization problems. For example, Gurobi optimization takes around 900s to solve one instance of TSP20 while GTA-RL ran with the same computational resources only takes around \emph{0.45 seconds} (the inference time) with beam search and \emph{0.39 seconds} with the greedy strategy to solve a batch of 100 problem instances. The inference time of GTA-RL and other learning baselines are similar (in between 0.3-0.5 seconds).

\textbf{Dynamic TSP (10,50):}
Compared to Figure \ref{fig:tsp20}, we observe a similar trend in Figure \ref{fig:tsp10} and \ref{fig:tsp50}.
In Figure \ref{fig:tsp10}, the learning-based algorithms (S2V-DQN-last, RNN-RL, AM) are able to find a reasonable solution, however, their performance degrades severely as we increase the number of nodes, as evident from Figure \ref{fig:tsp50}. This is because when the number of nodes increases, the tour length is longer, and the possibility of node location changes is higher. Due to the increase in dynamicity of node locations, S2V-DQN-last, RNN-RL, AM are not able to find good solutions for larger graphs. 

We achieve a similar optimal gap (of 8\%) in dynamic TSP10, demonstrating that the number of nodes does not affect much the optimal gap. Note however that we were unable to compute the optimal solution for dynamic TSP50 using the Gurobi optimization, due to its computational complexity, which reduces its applicability to larger real-life graphs.

\textbf{Dynamic VRP (10,20,50): } For Dynamic VRP, we observe a similar trend to TSP discussed in the previous section~\footnote{\textcolor{darkgreen}{Note that there is a slight performance increase in AM-D compared to the AM algorithm due to the dynamic encoder.}}. In Figure \ref{fig:vrp10}, \ref{fig:vrp20} and \ref{fig:vrp50}\footnote{We use pre-trained weights from dynamic VRP20 while training the GTA-RL and AM-D for VRP50}, GTA-RL-bs(10) finds the shortest tour length compared to the baselines and other GTA-RL variations. All the GTA-RL variations outperform the other learning-based baselines, and the performance of GTA-RL-rt is on par with GTA-RL-greedy. The results from GTA-RL variations demonstrate that our proposed architecture can handle both dynamic TSP and dynamic VRP.

We should also note that the optimal gap (of around 12\%) is similar to that of TSP, being slightly higher due to the increased complexity of VRP problem.
\pgfplotstableread[col sep = comma]{Tex/run-icde_dynamic_tsp_20_run_9-tag-val_avg_reward.csv}\dataNine
\pgfplotstableread[col sep = comma]{Tex/run-icde_dynamic_tsp_20_run_10-tag-val_avg_reward.csv}\dataTen
\pgfplotstableread[col sep = comma]{Tex/run-icde_dynamic_tsp_20_run_11-tag-val_avg_reward.csv}\dataElv
\pgfplotstableread[col sep = comma]{Tex/run-icde_dynamic_tsp_20_run_12-tag-val_avg_reward.csv}\dataTwe
\pgfplotstableread[col sep = comma]{Tex/run-icde_dynamic_tsp_20_run_14-tag-val_avg_reward.csv}\dataFteen

\pgfplotstableread[col sep = comma]{Tex/run-icde_dynamic_tsp_20_run_9-tag-avg_cost.csv}\dataNineTrain
\pgfplotstableread[col sep = comma]{Tex/run-icde_dynamic_tsp_20_run_10-tag-avg_cost.csv}\dataTenTrain
\pgfplotstableread[col sep = comma]{Tex/run-icde_dynamic_tsp_20_run_11-tag-avg_cost.csv}\dataElvTrain
\pgfplotstableread[col sep = comma]{Tex/run-icde_dynamic_tsp_20_run_12-tag-avg_cost.csv}\dataTweTrain
\pgfplotstableread[col sep = comma]{Tex/run-icde_dynamic_tsp_20_run_14-tag-avg_cost.csv}\dataFteenTrain

\begin{figure}[t]
\begin{subfigure}{0.5\columnwidth}
\begin{tikzpicture}
 \begin{axis}[
width=8.5cm, height=4.8cm,     
x tick label style={rotate=90},
xlabel=Epoch Number,
ylabel=Tour Length,
ylabel style={at={(0.1,0.5)}},
legend columns=3,
legend style={at={(0.5,0.7)},anchor=north, , font=\scriptsize, legend image post style={xscale=0.5}},
cycle list name=color list]

\addplot+[] table[x=Step, y, y=Value, col sep=comma]{\dataNine};
\addplot+[] table[x=Step, y, y=Value, col sep=comma]{\dataTen};
\addplot+[] table[x=Step, y, y=Value, col sep=comma]{\dataElv};
\addplot+[green] table[x=Step, y, y=Value, col sep=comma]{\dataTwe};
\addplot+[] table[x=Step, y, y=Value, col sep=comma]{\dataFteen};

\legend{GTA-RL-greedy, GTA-RL-sum, GTA-RL-(0), AM, GTA-RL-rt}
\centering
\end{axis}
\end{tikzpicture}
\caption{Tour length for validation data set during the training}
\label{fig:val}
\vspace{-2mm}
\end{subfigure}\begin{subfigure}{0.5\columnwidth}
\begin{tikzpicture}
 \begin{axis}[
width=8.5cm, height=4.8cm,     
x tick label style={rotate=90},
xlabel=Training Step,
ymax=8,
ylabel=Tour Length,
ylabel style={at={(0.1,0.5)}},
legend columns=3,
legend style={at={(0.54,0.98)},anchor=north, , font=\scriptsize, legend image post style={xscale=0.5}},
cycle list name=color list]

\addplot+[] table[x=Step, y, y=Value, col sep=comma]{\dataNineTrain};
\addplot+[] table[x=Step, y, y=Value, col sep=comma]{\dataTenTrain};
\addplot+[] table[x=Step, y, y=Value, col sep=comma]{\dataElvTrain};
\addplot+[green] table[x=Step, y, y=Value, col sep=comma]{\dataTweTrain};
\addplot+[] table[x=Step, y, y=Value, col sep=comma]{\dataFteenTrain};

\legend{GTA-RL-greedy, GTA-RL-sum, GTA-RL-(0), AM, GTA-RL-rt}
\centering
\end{axis}
\end{tikzpicture}
\caption{Tour length for training data set during the training}
\label{fig:trn}
\end{subfigure}
\end{figure}

\subsection{GTA-RL Variations Comparison} \label{sec-variations}

In this section, we first demonstrate the impact of our temporal encoder and temporal pointer. Figure \ref{fig:val} shows the average tour length achieved by each variation of GTA-RL solutions after each training epoch. We also show the AM approach on the graph, which is somewhat similar to our approach but does not use any temporal components. 

In Figure \ref{fig:val}, all the variations of GTA-RL achieve a lower tour length compared to the AM solution mainly because of the temporal encoder's ability to encode the temporal information. Next, even though GTA-RL-greedy, GTA-RL-sum, GTA-RL-(0) all use the same encoder, the decoder is different. As we explained, GTA-RL-sum and GTA-RL-(0) do not use the temporal decoder and naively compute a fixed embedding from the encoding output. These two are not able to achieve good performance compared to GTA-RL-greedy, where the temporal decoder is used, which shows the benefit of using the temporal decoder. We should also note that GTA-RL-(0) is  performing somewhat better compared to GTA-RL-sum because, in GTA-RL-sum, all the encoder output is summed up in the temporal axis, which loses information about dynamic changes. However, in GTA-RL-(0), even though it uses a single time step data, GTA-RL-(0) will preserve  temporal data up to a certain extent. 

Next, we use both Figure \ref{fig:val} and Figure \ref{fig:trn} to show how close the GTA-RL-rt and GTA-RL-greedy solutions are. In Figure \ref{fig:val}, compared with the other solutions, GTA-RL-rt achieves  similar performance to GTA-RL-greedy. This demonstrates that even in real-time settings, our proposed method can find a high-quality solution. As shown in Figure \ref{fig:trn}, the tour length during the training of GTA-RL-rt is also much closer to GTA-RL-greedy compared to other solutions. However, it is interesting to see that there is a variation of GTA-RL-rt that is higher than what is found by GTA-RL-greedy. This is mainly because GTA-RL-rt, at its first-time step, takes the decision only based on the current time step node locations. Since node features are changing uniformly at random, GTA-RL-rt cannot guarantee that the selected node will be the best decision given that node features can change externally in future. This randomness can cause a volatility of the tour length. Despite these, the average tour of GTA-RL-rt is able to be  close to GTA-RL-greedy.

\subsection{Generalization} \label{sec-gen}

In this section, we show how GTA-RL model can generalize to larger graphs beyond those used for training. To find such a generalization, we first compute the tour length for a set of graphs with 50 nodes by GTA-RL trained originally with a set of graphs with 20 nodes. Then, for comparison, we compute the tour length for the same set of graphs originally trained for graphs with 50 nodes. These results are shown in Table \ref{tbl:gen}. In both situations (in dynamic TSP50 and in dynamic VRP50), the tour length achieved by GTA-RL trained with 20 nodes is close to GTA-RL trained with 50 nodes with only around 3\% of discrepancy. This shows that GTA-RL can generalize for much larger graphs than originally trained for. 

\begin{table}
\begin{tabular}{|c|c|c|c|}
\hline
Problem       & GTA-RL trained with 50 nodes & GTA-RL trained with 20 nodes & Gap   \\ \hline
Dynamic TSP50 & 8.17                                                                    & 8.48                                                                    & 3.6\% \\ \hline
Dynamic VRP50 & 13.12                                                                   & 13.71                                                                    & 4.3\% \\ \hline
\end{tabular}
\caption{Tour length of Dynamic TSP50 and Dynamic VRP50 from GTA-RL trained with networks with 20 and 50 nodes. Beam width of 10 is used.}
\label{tbl:gen}
\centering
\end{table}

\subsection{Visualizing the Dynamic TSP}
\begin{figure*}
\centering
\begin{subfigure}[t]{.35\textwidth}
\centering
    \includegraphics[width=5cm, height=3.6cm]{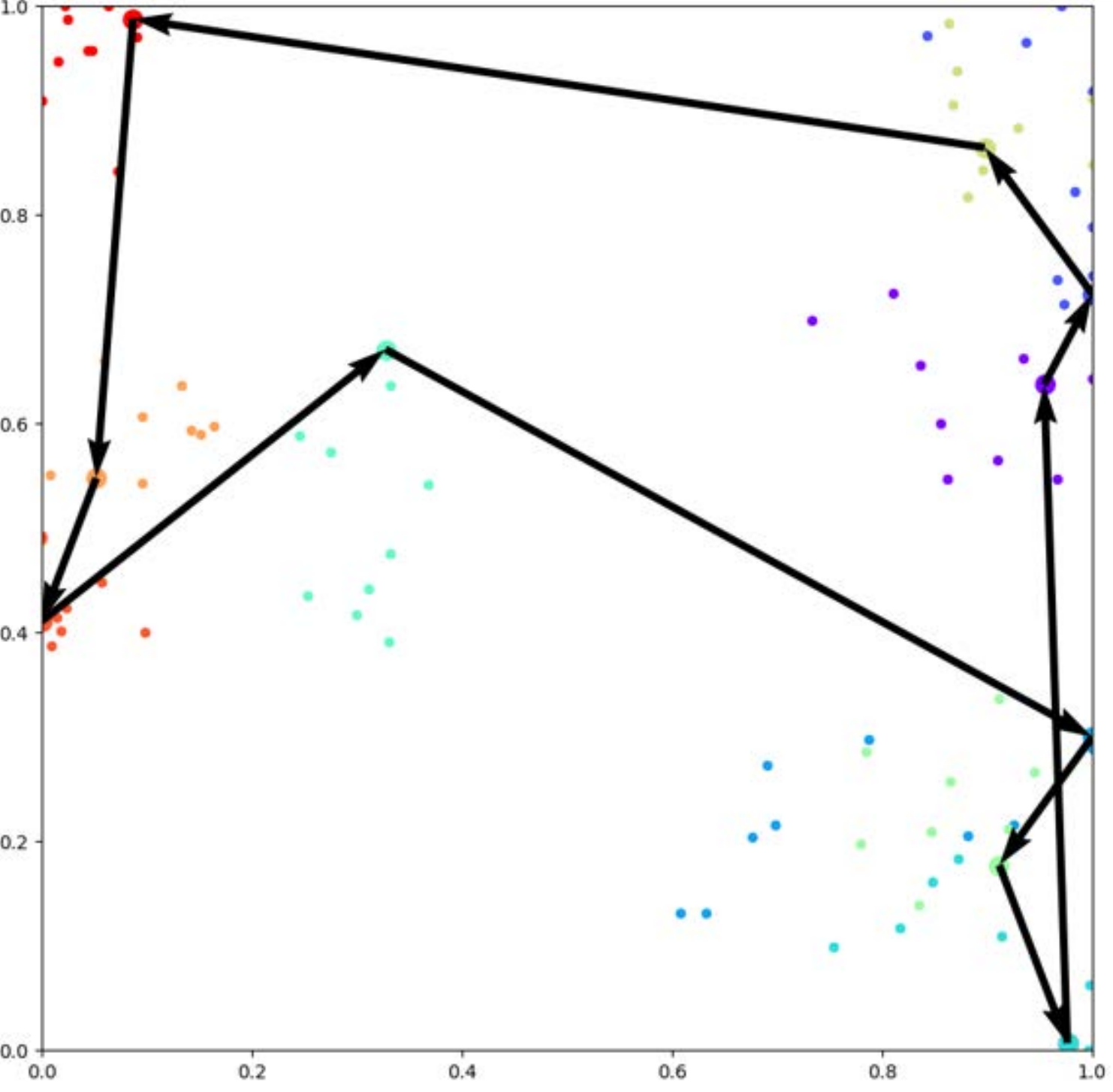}
    \caption{AM with a tour length of 3.85}
    \label{fig:vis_a}
\end{subfigure}\begin{subfigure}[t]{.33\textwidth}
\centering
    \includegraphics[width=5cm, height=3.6cm]{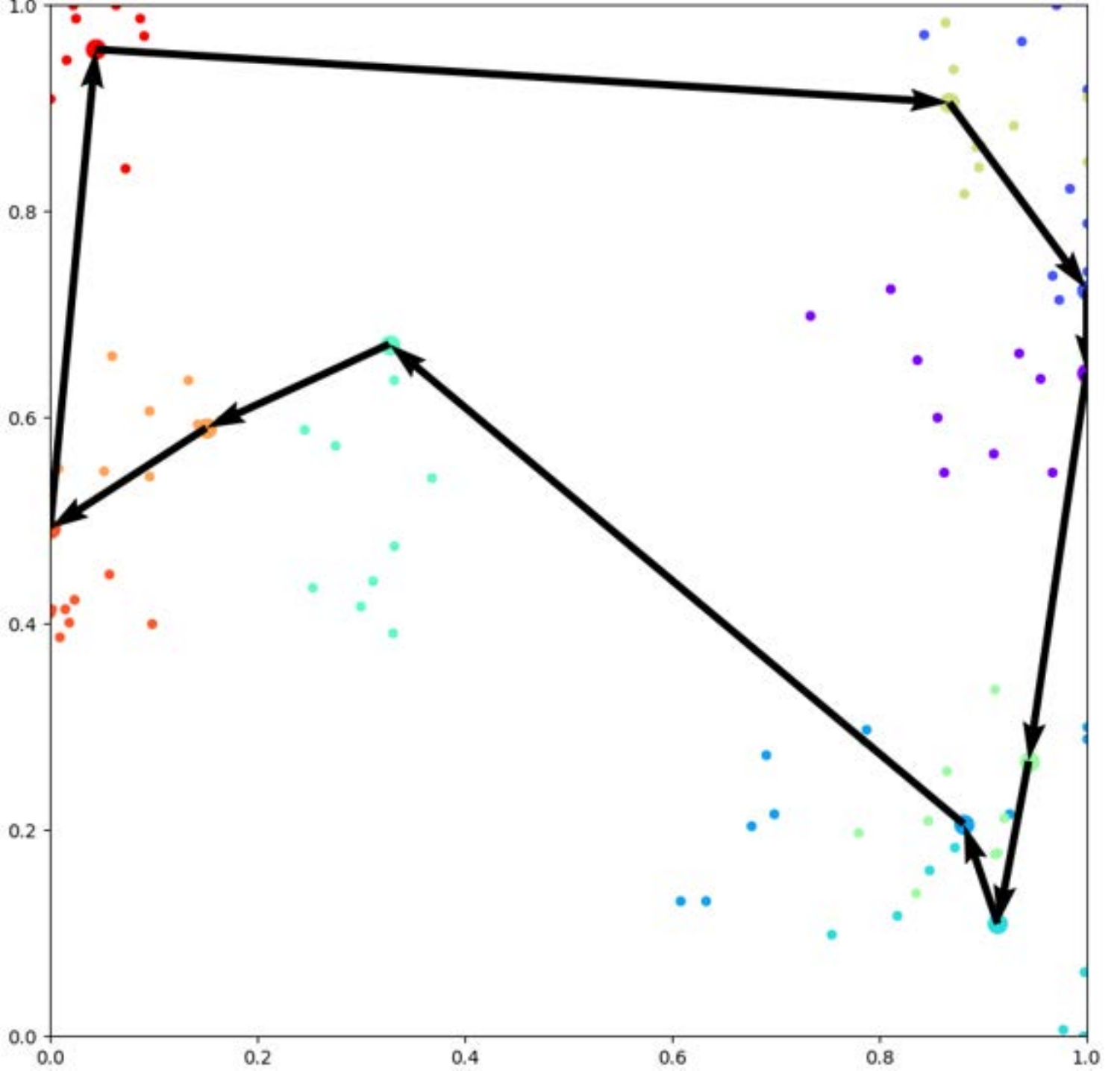}
        \caption{GTA-RL-greedy with a tour length of 3.15}
            \label{fig:vis_b}
\end{subfigure}\begin{subfigure}[t]{.33\textwidth}
\centering
    \includegraphics[width=5cm, height=3.6cm]{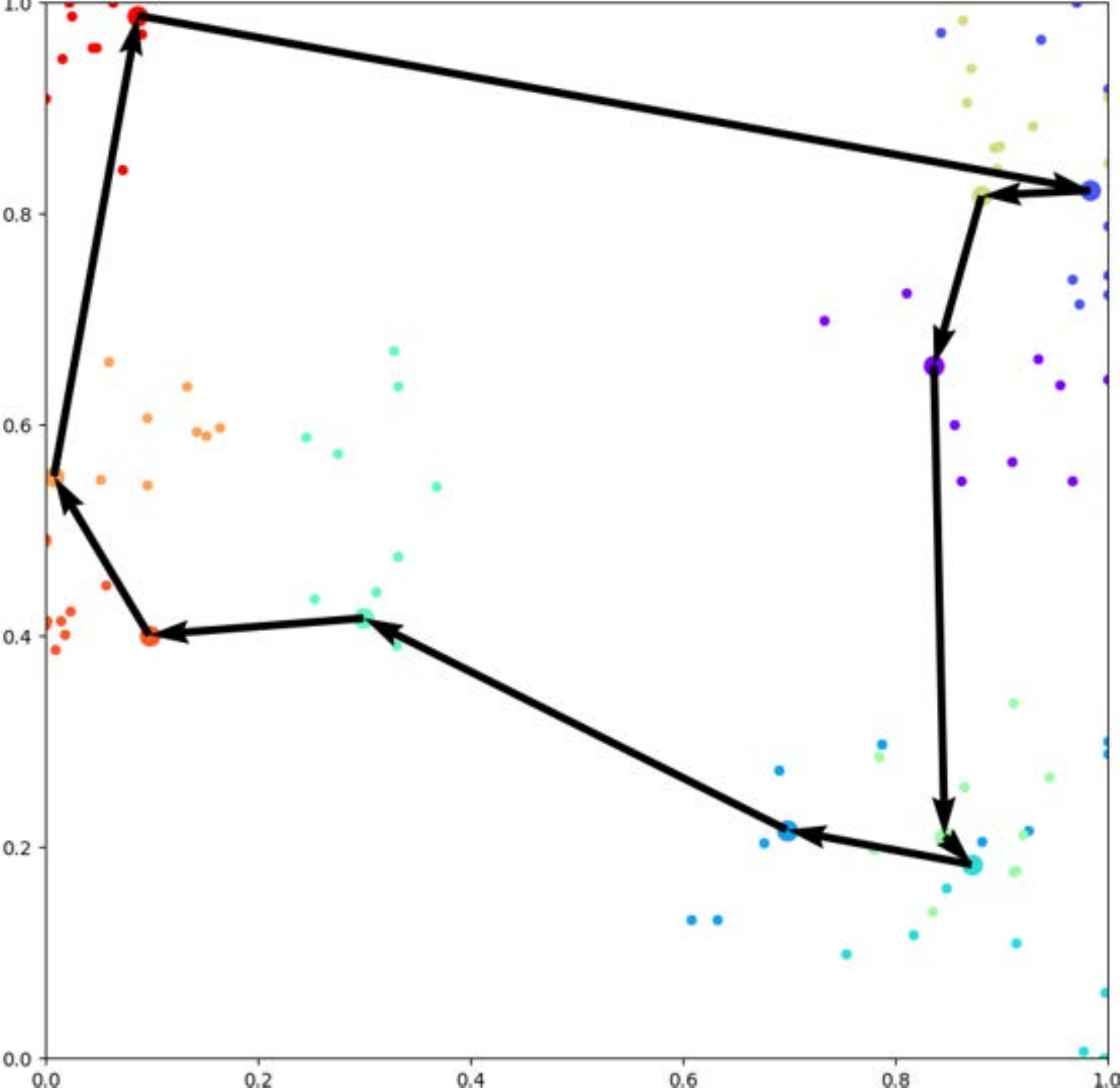}
        \caption{Gurobi solution with a tour length 2.96}
            \label{fig:vis_c}
\end{subfigure}
\caption{Figure shows the order of selected nodes by AM, GTA-RL-greedy and Gurobi optimal solver. For easy visualization we use dynamic TSP10 where only 10 cities are present. The dots with the same color indicates the same city locations at different time steps. The black arrows shows the tour decision taken by each algorithm.}
\label{fig:visualize}
\end{figure*}
In this experiment, we visualize the tours computed by AM, GTA-RL-greedy and Gurobi optimal solver for dynamic TSP 10 in Figure \ref{fig:visualize}. The set of nodes with the same color represents the locations of the same node in different time steps.
Figure \ref{fig:vis_a} shows the tour output from AM is 3.86. In Figure \ref{fig:vis_a}, in the bottom right corner, AM is not able to find the exact order for traversal as nodes which are further were visited before the closer ones. GTA-RL-greedy (presented in Figure \ref{fig:vis_b}) is able to avoid that and select the order of nodes to achieve a shorter tour length of 3.15, which is much closer to the optimal solution found by Gurobi presented in Figure \ref{fig:vis_c}. 

\section{Conclusion}

In this work, we propose an effective and efficient learning-based method to solve dynamic graph combinatorial optimization problems, which is crucial when applying combinatorial optimization to real-world applications. Our proposed temporal encoder-decoder architecture is able solve dynamic graph problems by first, learning a representation for the given problem instance and second, dynamically focusing on different parts of the learned representation to find a solution sequence to the problem instance. We then introduce a variant of our proposed model to solve real-time combinatorial optimization where input features are known only after making a decision. We evaluate our method by comparing it against several state-of-the-art approaches, both learning-based methods, as well as hand-crafted heuristics, and against the optimal controllers, and show that \textcolor{darkgreen}{our approach substantially outperforms both learning-based and hand-crafted heuristics and is on par with the optimal controller when solving famous NP-hard problems (TSP and VRP) in transportation.}

There are several interesting directions of future work. First, scalability is an open problem in the neural CO domain and Section \ref{sec-gen} provides promising results towards this direction and can be investigated further. Second, our real-time encoder encodes at every time step by buffering the previous states which can be memory expensive and this can be investigated further to store and process data more efficiently. Third, this approach can be applied to many real-world applications such as flow optimization in traffic engineering, determining autonomous truck routes, bandwidth allocations for telecommunications networks \textcolor{darkgreen}{and for knowledge graph reasoning in graph databases}. 
Finally, dynamic combinatorial optimization is much harder than static combinatorial optimization; thus, proposing learning heuristics that are close to the optimal is much more challenging. As this is a first step in learning heuristics for dynamic graph combinatorial optimization, we hope to reduce the gap between our proposed approach and the optimal controller in future research.

\bibliographystyle{unsrt}  
\bibliography{references}

\end{document}